\documentclass[10pt,twocolumn,letterpaper]{article}

\usepackage{cvpr}              

\definecolor{cvprblue}{rgb}{0.21,0.49,0.74}
\usepackage[pagebackref,breaklinks,colorlinks,allcolors=cvprblue]{hyperref}

\def\paperID{3494} 
\def\confName{CVPR}
\def\confYear{2025}

\title{APHQ-ViT: Post-Training Quantization with Average Perturbation Hessian Based Reconstruction for Vision Transformers}

\author{
    Zhuguanyu Wu$^{1,2}$, 
    Jiayi Zhang$^{1,2}$,  
    Jiaxin Chen$^{1,2}$\textsuperscript{\Letter}, 
    Jinyang Guo$^{3}$, 
    Di Huang$^{2}$, 
    Yunhong Wang$^{1,2}$\textsuperscript{\Letter} \\
    $^1$State Key Laboratory of Virtual Reality Technology and Systems, Beihang University, China \\
    $^2$School of Computer Science and Engineering, Beihang University, Beijing, China \\
    $^3$School of Artificial Intelligence, Beihang University, Beijing, China \\
    {\tt\small \{goatwu, zhangjyi, jiaxinchen, jinyangguo, dhuang, yhwang\}@buaa.edu.cn}
}

\usepackage{bm}
\usepackage{amsthm}
\usepackage{algorithm}
\usepackage{algpseudocode}
\usepackage{marvosym}
\usepackage{graphicx}	
\usepackage{amsmath}	
\usepackage{amssymb}	
\usepackage{booktabs}
\usepackage{times}
\usepackage{microtype}
\usepackage{epsfig}
\usepackage{caption}
\usepackage{float}
\usepackage{placeins}
\usepackage{color, colortbl}
\usepackage{stfloats}
\usepackage{enumitem}
\usepackage{tabularx}
\usepackage{xstring}
\usepackage{multirow}
\usepackage{xspace}
\usepackage{url}
\usepackage{subcaption}
\usepackage[hang,flushmargin]{footmisc}

\newtheorem{theory}{Theorem}[section]

\begin{document}
\maketitle
\begin{abstract}
Vision Transformers (ViTs) have become one of the most commonly used backbones for vision tasks. Despite their remarkable performance, they often suffer significant accuracy drops when quantized for practical deployment, particularly by post-training quantization (PTQ) under ultra-low bits. Recently, reconstruction-based PTQ methods have shown promising performance in quantizing Convolutional Neural Networks (CNNs). However, they fail when applied to ViTs, primarily due to the inaccurate estimation of output importance and the substantial accuracy degradation in quantizing post-GELU activations. To address these issues, we propose \textbf{APHQ-ViT}, a novel PTQ approach based on importance estimation with Average Perturbation Hessian (APH). Specifically, we first thoroughly analyze the current approximation approaches with Hessian loss, and propose an improved average perturbation Hessian loss. To deal with the quantization of the post-GELU activations, we design an MLP Reconstruction (MR) method by replacing the GELU function in MLP with ReLU and reconstructing it by the APH loss on a small unlabeled calibration set. Extensive experiments demonstrate that APHQ-ViT using linear quantizers outperforms existing PTQ methods by substantial margins in 3-bit and 4-bit across different vision tasks. The source code is available at \url{https://github.com/GoatWu/APHQ-ViT}.
\end{abstract}
\renewcommand{\thefootnote}{}
\footnote{\textsuperscript{\Letter} Corresponding Authors}
\renewcommand{\thefootnote}{\arabic{footnote}}
\addtocounter{footnote}{-1}
\section{Introduction}
\label{sec:intro}

\begin{figure}[t]
    \centering
    \begin{subfigure}{1.\columnwidth}
        \centering
        \includegraphics[width=1\linewidth]{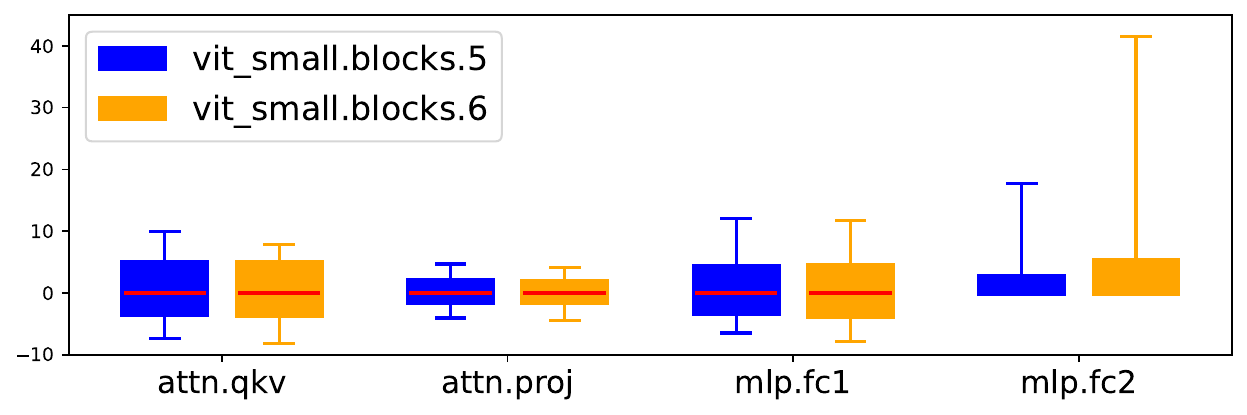}
        \caption{activation statistics of linear layers}
        \label{fig:fc2_sub3}
    \end{subfigure}%
    \hfill
    \begin{subfigure}{.50\columnwidth}
        \centering
        \includegraphics[width=1\linewidth]{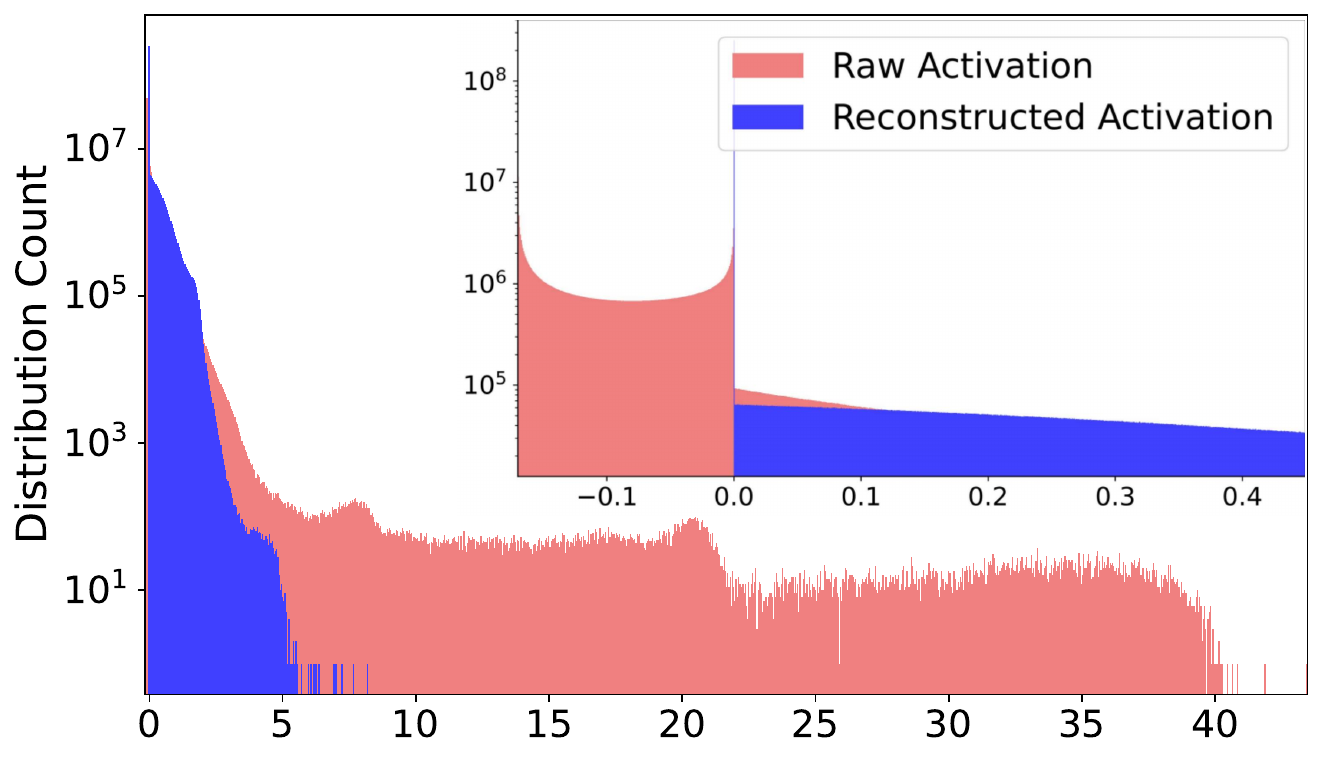}
        \caption{activation distribution of \textit{mlp.fc2}}
        \label{fig:fc2_sub1}
    \end{subfigure}%
    \hfill
    \begin{subfigure}{.50\columnwidth}
        \centering
        \includegraphics[width=1\linewidth]{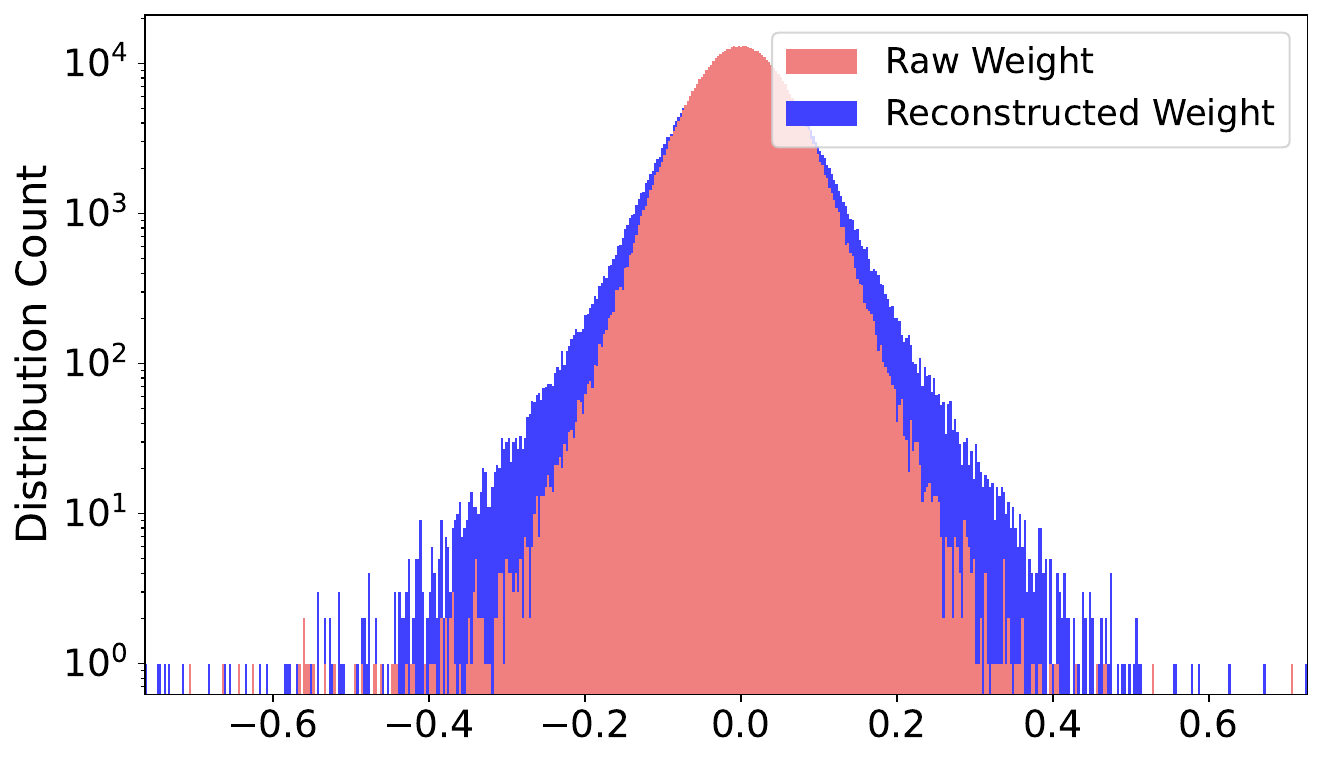}
        \caption{weight distribution of \textit{mlp.fc2}}
        \label{fig:fc2_sub2}
    \end{subfigure}
    \caption{(a) shows the box plot of activations for each linear layer in \textit{vit-small.blocks.5 and 6} by using the 0.99 quantile, highlighting the varying ranges of post-GELU activations. (b) and (c) display that the activation range of \textit{fc2} is significantly reduced after MLP Reconstruction, while the weight range only exhibits slight changes.}
    \label{fig:fc2_dist}
\end{figure}

The success of Transformer-based models in natural language processing (NLP) \cite{attention,bert} has inspired their application to various computer vision tasks, such as image classification \cite{vit, swin, crossvit}, object detection \cite{objdettrans, detr, dydetr, yewen} and instance segmentation \cite{segmenter, segsts, effiseg, octr, catdet}. Due to their sophisticated architectures for representation learning, substantial memory usage and computational overhead make it a great challenge to deploy these models on resource-constrained devices~\cite{whitepaper}.

Model quantization has recently emerged as a promising solution to reduce the computational cost of deep learning models. This technique converts the weights or activations from float-point precision to low bit-width, while preserving the original model architectures. Most current quantization approaches are generally categorized into two groups: quantization-aware training (QAT) \cite{pact,stepsize} and post-training quantization (PTQ) \cite{quantsurvey2,quantsurvey}. QAT methods typically achieve superior accuracy compared to PTQ by performing end-to-end training on the full pretraining dataset. Nevertheless, they are often time-intensive and encounter substantial limitations when the original dataset is inaccessible. In contrast, PTQ methods are more applicable as they rely solely on a small unlabeled calibration dataset instead of requiring access to the full training set. PTQ methods can be further divided into two categories, \ie, the one that only involves calibration \cite{FQViT,PTQ4ViT,RepQViT,ptq4sam}, and the reconstruction-based one \cite{isvit,dopq,oasq,haocheng}, where the later generally achieves superior accuracy by introducing an efficient fine-tuning process. Despite their promising performance in quantizing CNNs~\cite{BRECQ,qdrop,pdquant}, the reconstruction-based methods suffer from the following two limitations when applied to ViTs.

1) \emph{Inaccurate estimation of output importance.} Representative reconstruction-based PTQ methods employ the block-reconstruction framework \cite{BRECQ,qdrop}, which fine-tunes the AdaRound \cite{adaround} weights to ensure that the output of the quantized block closely matches the output of the original full-precision one. The mean squared error (MSE) between the quantized and original outputs is one of the most commonly used metrics to evaluate quantization quality. However, this approach is suboptimal, since it treats all output tokens and dimensions equally, overlooking the critical importance of the class token and the importance variations across channels in ViTs as shown in Sec. \ref{sec:visualize} of the \emph{supplementary material}. Some works leverage the Hessian matrix based on Fisher information to explore the distinct importance~\cite{BRECQ,PTQ4ViT,APQViT}, while fail to surpass the MSE loss due to the inaccurate approximation on the Hessian matrix. 

2) \emph{Performance degradation in quantizing post-GELU activation.} As shown in \cref{fig:fc2_dist} (a), the quantization error in post-GELU activations stems from two primary factors. First, the activation distribution is highly imbalanced: negative activations are densely concentrated within a narrow interval \([-0.17,0]\), while positive activations follow sparse distributions. Second, the activation range varies significantly, reaching up to 40 in certain layers. Some works have attempted to deal with the imbalanced activation distribution by a twin-uniform quantizer that employs separate scaling factors for positive and negative activations \cite{PTQ4ViT}, or employ a hardware-friendly logarithmic quantizer with an arbitrary base \cite{adalog}. However, they necessitate specialized hardware support for the quantizer, limiting their practicality in real-world applications.

To address the above issues, we propose a novel quantization approach dubbed APHQ-ViT for the post-training quantization of Vision Transformers. As illustrated in \cref{fig:overview}, to tackle the \emph{inaccurate estimation of output importance}, we thoroughly investigate the current approximation methods with Hessian loss, and propose an improved average perturbation Hessian (APH) loss for block reconstruction. We show that applying APH to explore the importance of output can stabilize the reconstruction process, and further promote precision. To deal with \emph{performance degradation in quantizing post-GELU activations}, we develop the MLP Reconstruction method (MR), by replacing the GELU activation function with ReLU. As shown in \cref{fig:fc2_dist}(b) and (c), MR not only reduces the activation range while maintaining the weight range, but also alleviates the imbalanced activation distributions, thus reducing the quantization error.

The main contributions of our work lie in three-fold:

1) We thoroughly analyze the limitations of existing Hessian guide quantization loss, and propose an improved Average Perturbation Hessian (APH) loss by mitigating the estimation deviations, which facilitates both the block-wise quantization reconstruction and MLP reconstruction.

2) We develop a novel MLP Reconstruction (MR) method by replacing the GELU activation function in MLP with ReLU, which simultaneously alleviates imbalanced activation distribution and significantly reduces the activation range, making the model more amenable to quantization.

3) We extensively conduct experiments on public datasets across various vision tasks in order to evaluate the performance of our method. Experimental results demonstrate that the proposed method, utilizing only linear quantizers, significantly outperforms the current state-of-the-art approaches with distinct Vision Transformer architectures, especially in the case of ultra-low bit quantization.

\section{Related Work}
\label{sec:related}

Model quantization, which aims to map the floating-point weights and activations to lower bit widths, has become one of the most widely used techniques for accelerating the inference of deep learning models. It can be roughly divided into two categories: Post-Training Quantization (PTQ) and Quantization Aware Training (QAT). Among the quantization methods for Vision Transformers, QAT methods \cite{pact,stepsize,bivit,qvit} often achieve higher accuracy. However, QAT methods often require a large amount of training resources, limiting their universality. By contrast, PTQ methods only take a small calibration dataset to adjust quantization parameters, making them resource-efficient.

The PTQ methods can be further categorized into two groups, \emph{i.e.}, the calibration-only methods that solely involve the calibration stage, and the reconstruction-based methods that additionally incorporate a reconstruction stage.

Calibration-only methods can efficiently obtain a quantized model. PTQ4ViT \cite{PTQ4ViT} employs a twin-uniform quantizer to reduce the activation quantization error, and adopts a Hessian guided loss to evaluate the effectiveness of different scaling factors. RepQ-ViT \cite{RepQViT} decouples the quantization and inference processes, specifically addressing post-LayerNorm activations with significant inter-channel variations. NoisyQuant \cite{noisyquant} reduces quantization error by adding a fixed uniform noisy bias to the values being quantized. IGQ-ViT \cite{igqvit} employs a group-wise activation quantizer to balance the inference efficiency and quantization accuracy. ERQ \cite{erq} introduces the GPTQ approach \cite{gptq} to ViTs and proposes an activation quantization error reduction module to mitigate quantization errors, along with a derived proxy for output error to refine weight rounding. AdaLog \cite{adalog} designs a hardware-friendly arbitrary-base logarithmic quantizer to handle power-law activations and a progressive hyperparameter search algorithm. However, these methods still suffer substantial quantization loss under low-bit quantization.

Reconstruction-based methods often achieve quantized models with higher accuracy, by additionally employing quantization reconstruction. Numerous approaches have been developed for CNNs. AdaRound \cite{adaround} adopts a refined weight rounding strategy to minimize the task loss, outperforming conventional rounding-to-nearest methods. BRECQ \cite{BRECQ} improves performance by leveraging cross-layer dependencies through block-wise reconstruction. QDrop \cite{qdrop} employs random activation dropout during block reconstruction, facilitating obtaining smoother optimized weight distributions. Although effective for CNNs, these methods yield suboptimal results when applied to ViTs. I\&S-ViT \cite{isvit} employs a three-stage smooth optimization strategy to address the quantization inefficiency and ensure stable learning. DopQ-ViT selects optimal scaling factors to mitigate the impact of outliers and preserve quantization performance. OASQ \cite{oasq} addresses outlier activations employing distinct granularities in the quantization reconstruction. Although these methods generally outperform calibration-only approaches, they still struggle to reach an acceptable performance under ultra-low bit quantization.

\section{The Proposed Approach} 

\begin{figure*}[t]
    \centering
    \includegraphics[width=1\linewidth]{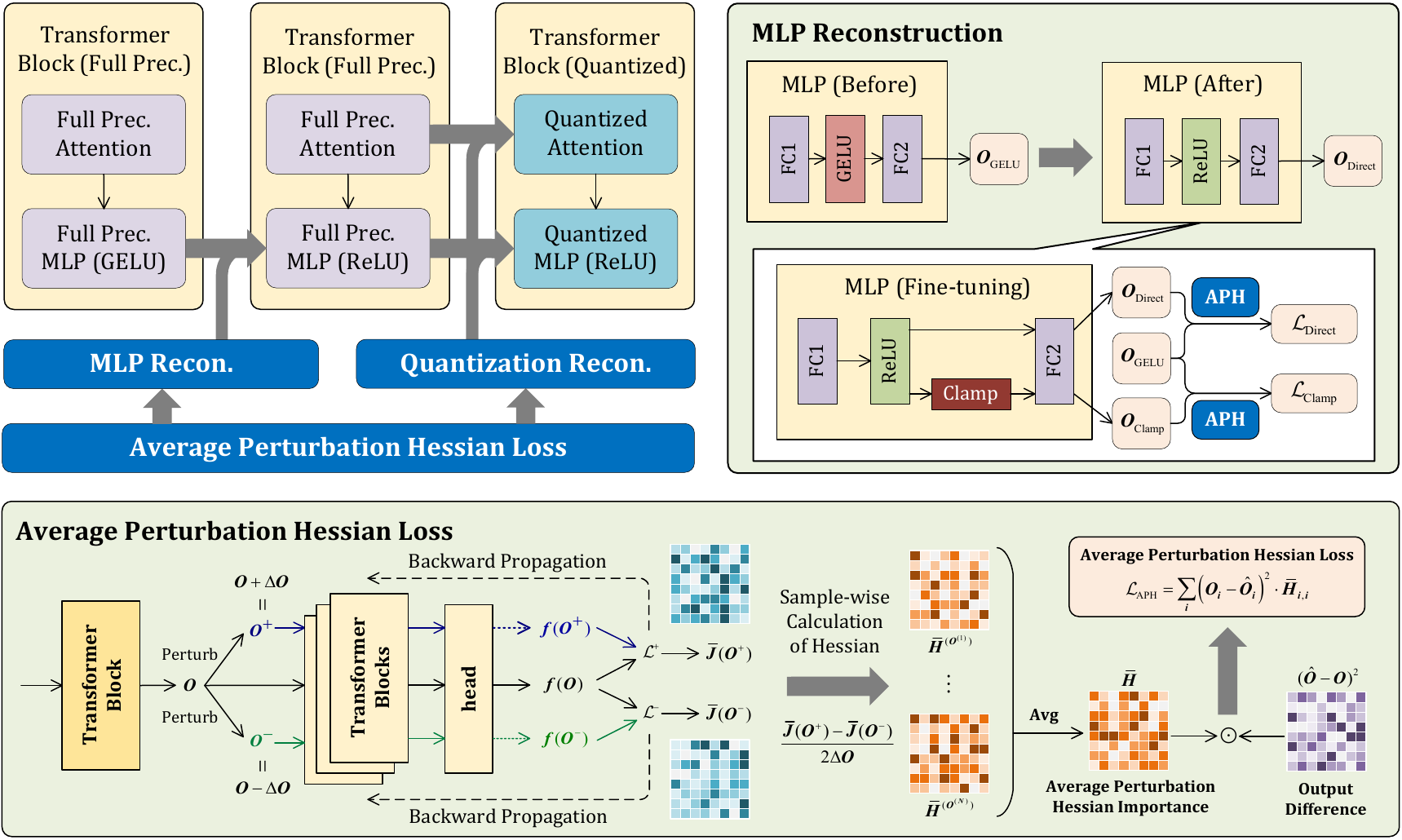}
    \caption{Framework overview of APHQ-ViT. In the block-wise quantization process, we first reconstruct the MLP layer, followed by quantization reconstruction, both of which are optimized by the proposed Average Perturbation Hessian (APH) loss. The MLP Reconstruction (MR) method replaces the GELU activation function with ReLU and reduces the post-GELU activation range. The detailed implementation of the APH loss is visualized at the bottom. }
    \label{fig:overview}
\end{figure*}

\noindent As shown in \cref{fig:overview}, the proposed APHQ-ViT approach follows the block-wise quantization pipeline. In each block, we first perform MLP Reconstruction, followed by quantization reconstruction based on QDrop. The average perturbation Hessian loss is applied in both reconstructions to explore the distinct output importance. The overall pipeline of APHQ-ViT is summarized in Algorithm \ref{alg:overview}. The average perturbation Hessian loss and MLP Reconstruction are described in \cref{sec:aph} and \cref{sec:mlp_recon}, respectively.

\begin{algorithm}[t]
\caption{APHQ-ViT for Block-wise Quantization.}
\label{alg:overview}
\begin{algorithmic}[1]
    \Statex \textbf{Input:} The full-precision model \(\mathcal{M}\), the full-precision block \(\mathcal{B}\) to be quantized, the calibration data \(\mathcal{D}_{calib}\), and the loss function \(\mathcal{L}\).
    \Statex \noindent \textcolor[RGB]{82,147,141}{ \# Calculate the Average Perturbation Hessian: }
    \State Compute the raw output \(\bm{O}\) of \(\mathcal{B}\) based on \(\mathcal{D}_{calib}\).
    \State Calculate the perturbed outputs \(\bm{O}^+\) and \(\bm{O}^-\).
    \State Compute \(f(\bm{O})\)/\(f(\bm{O}^+)\)/\(f(\bm{O}^-)\) by forward passing \(\bm{O}\)/\(\bm{O}^+\)/\(\bm{O}^-\) through the remaining blocks of \(\mathcal{M}\).
    \State Calculate \(\mathcal{L}(f(\bm{O}), f(\bm{O}^+))\) and \(\mathcal{L}(f(\bm{O}), f(\bm{O}^-))\) and obtain \(\bar{\bm{J}}^{(\bm{O}^+)}\) and \(\bar{\bm{J}}^{(\bm{O}^-)}\) by backward propagation.
    \State Calculate the average perturbation Hessian matrix \(\bar{\bm{H}}\) based on Eq.~\eqref{eq:aph}.
    \Statex \noindent \textcolor[RGB]{82,147,141}{ \# MLP Reconstruction: }
    \State Replace the GELU activation of MLP by ReLU.
    \For{$i = 0$, $\cdots$, \(\mathrm{max\_iter}\)} 
        \State Calculate \(\bm{O}_{\mathrm{Direct}}\) and \(\bm{O}_{\mathrm{Clamp}}\) by Eqs. \eqref{eq:o_direct} and \eqref{eq:mlprecon}.
        \State Calculate \(\mathcal{L}_{\mathrm{Distill}}\) by Eq.~\eqref{eq:l_distill}.
        \State Perform backward propagation and update MLP.
    \EndFor
    \Statex \noindent \textcolor[RGB]{82,147,141}{ \# Quantization Reconstruction: }
    \For{$i = 0$, $\cdots$, \(\mathrm{max\_iter}\)}
        \State Calculate the quantized output \(\widehat{O}\) by QDrop \cite{qdrop}.
        \State Calculate \(\mathcal{L}_{\mathrm{APH}}\) based on Eq.~\eqref{eq:l_aph}.
        \State \parbox[t]{\dimexpr\linewidth-\algorithmicindent}{Perform backward propagation and update the AdaRound \cite{adaround} weights in \(\mathcal{B}\).}
    \EndFor
    \Statex \textbf{Output:} The quantized block \(\widehat{\mathcal{B}}\).
\end{algorithmic}
\end{algorithm}

\subsection{Preliminaries: Hessian in BRECQ} \label{sec:hessian_pre}

The Hessian guided metric proposed by BRECQ \cite{BRECQ} stands out as one of the most prevalent metrics for evaluating the quantization quality of CNNs. It assumes that the de-quantized weight $\widehat{\bm{W}}$ can be represented as the original weight $\bm{W}$ perturbed by $\bm{\epsilon}$, \emph{i.e.}, $\widehat{\bm{W}}=\bm{W}+\bm{\epsilon}$. The quality of quantization is measured by estimating the quantization loss through a Taylor expansion:
\begin{equation}\label{eq:taylor_1}
    \mathbb{E}[\mathcal{L}(\widehat{\bm{W}})] - \mathbb{E}[\mathcal{L}(\bm{W})] \approx \bm{\epsilon}^\top \bar{\bm{J}}^{(\bm{W})} + \frac{1}{2}\bm{\epsilon}^\top\bar{\bm{H}}^{(\bm{W})}\bm{\epsilon},
\end{equation}
where \(\bar{\bm{J}}^{(\bm{W})}\) and \(\bar{\bm{H}}^{(\bm{W})}\) are the Jacobian and Hessian matrices w.r.t the weight \(\bm{W}\), respectively.

Supposing the convergence of a pre-trained model to be quantized, existing works often drop the first-order term \(\bm{\epsilon}^\top \bar{\bm{J}}^{(W)}\) and approximate the Hessian matrix with squared gradient, resulting in the quantization loss:
\begin{equation}\label{eq:talor_appro_2}
    \mathbb{E}[\mathcal{L}(\widehat{\bm{W}})] - \mathbb{E}[\mathcal{L}(\bm{W})]
    \approx \sum_i\left((\widehat{\bm{O}}_i - \bm{O}_i) \cdot \frac{\partial \mathcal{L}}{\partial \bm{O}_i}\right)^2,
\end{equation}
where \(\bm{O}\) is the output, and \(\widehat{\bm{O}}\) is the de-quantized one of \(\bm{O}\).

The above Hessian guided quantization loss adopts two approximations as in BRECQ: 1) the Hessian matrix is approximated by the Fisher Information Matrix (FIM)~\cite{fisher1922mathematical}; 2) the diagonal elements of FIM are approximated by the squared gradients w.r.t. the output. These approximations achieve high accuracy, when the task loss is the Cross-Entropy (CE) loss, and the model's predicted distribution aligns closely with the true data distribution. However, in practice, models are often unable to fit the true data distribution well, leading to inevitable approximation errors. Additionally, these approximations fail to generalize to tasks such as segmentation and object detection. As a consequence, when applied to ViTs, the loss in Eq.~\eqref{eq:talor_appro_2} is inferior to the MSE Loss in many ViT architectures as shown in Table~\ref{tab:ab_hessian}. 

\subsection{Average Perturbation Hessian Loss} \label{sec:aph}

To address the limitations of the Hessian guided loss in Eq.~\eqref{eq:talor_appro_2}, we develop a perturbation based estimation method that only relies on two fundamental assumptions as below: 

A.1) When performing Taylor series expansions for the loss, the third and higher-order derivatives can be omitted without significantly sacrificing the accuracy \cite{adaround,kfac,hawq}.

A.2) The influence of individual elements on the final output is assumed to be independent, allowing the use of the diagonal Hessian as a practical substitute for the computationally intensive full Hessian \cite{hawq,lecun}. 

It is worth noting that A.1) and A.2) are widely used in various model compression methods, including BRECQ, which further rely on additional, stronger assumptions to achieve their results. 

We first extend the loss function to ensure compatibility across diverse tasks. Instead of the conventional CE loss, we regard quantization as a knowledge distillation process on a small unlabeled calibration dataset. This allows us to directly employ distillation loss to address different tasks. Specifically, for classification, we adopt the KL divergence between the output logits as the distillation loss. For two-stage object detection and instance segmentation, we combine the KL divergence from the classification head and the smooth L1 distance \cite{fastrcnn} from the regression head as the distillation loss.  Compared to the CE Loss, these distillation losses share the following common characteristics: \(\mathcal{L}(\hat{\bm{O}},\bm{O}) \ge 0\), and \(\mathcal{L}(\hat{\bm{O}},\bm{O}) = 0\) if and only if \(\bm{O} = \hat{\bm{O}}\). According to the extreme value theorem \cite{stewart2015calculus}, if \(\mathcal{L}\) is differentiable at $\hat{\bm{O}} = \bm{O}$, then we have:
\begin{equation} \label{eq:jacobian}
   \bar{\bm{J}}^{(\bm{O})} = \left. \frac{\partial \mathcal{L}(\hat{\bm{O}}, \bm{O})}{\partial \hat{\bm{O}}} \right|_{\hat{\bm{O}}=\bm{O}} = 0.
\end{equation}

Based on Eq.~\eqref{eq:jacobian}, we treat the errors introduced by quantization or MLP Reconstruction as small perturbations denoted by \(\bm{\epsilon}\), and perform a Taylor expansion as below:
\begin{align}
\begin{split}
    \mathcal{L}(\bm{O}+&\bm{\epsilon}) - \mathcal{L}(\bm{O}) = \bm{\epsilon}^\top \bar{\bm{J}}^{(\bm{O})} + \frac{1}{2}\bm{\epsilon}^\top\bar{\bm{H}}^{(\bm{O})}\bm{\epsilon} + O(\|\bm{\epsilon}\|^3) \\
    &= \frac{1}{2}\bm{\epsilon}^\top\bar{\bm{H}}^{(\bm{O})}\bm{\epsilon} + O(\|\bm{\epsilon}\|^3) \approx \frac{1}{2}\bm{\epsilon}^\top\bar{\bm{H}}^{(\bm{O})}\bm{\epsilon}, \\
\end{split}
\end{align}
where \(\mathcal{L}(\bm{O}+\bm{\epsilon}), \mathcal{L}(\bm{O})\) are the abbreviations of \(\mathcal{L}(\bm{O}+\bm{\epsilon},\bm{O})\) and \(\mathcal{L}(\bm{O},\bm{O})\), respectively, \(\bar{\bm{H}}^{(\bm{O})}\) is the Hessian matrix of \(\mathcal{L}\) w.r.t \(\bm{O}\), and \(O(\|\bm{\epsilon}\|^3)\) represents the sum of the third and higher-order derivatives. As depicted in \cref{eq:jacobian}, \(\mathcal{L}(\bm{O})\) and \(\bar{\bm{J}}^{(\bm{O})}\) are zeros, and \(O(\|\bm{\epsilon}\|^3)\) is omitted according to A.1).

Based on A.2), we follow BRECQ by utilizing the block-diagonal Hessian and disregarding the inter-block dependencies. By definition, the diagonal elements of the Hessian matrix are the second partial derivatives of the loss function:
\begin{equation}\label{eq:def_hessian}
    \bar{\bm{H}}^{(\bm{O})}_{i,i} = \frac{\partial^2 \mathcal{L}}{\partial \bm{O}^2_{i}} = \frac{\partial}{\bm{O}_{i}} \left(\frac{\partial \mathcal{L}}{\bm{O}_{i}}\right).
\end{equation}
For the $i-$th diagonal element, we perturb \(\bm{O}\) by \(\Delta\bm{O}=10^{-6}\): \(\bm{O}^+ = \bm{O} + \Delta\bm{O}\cdot \mathbf{1}\) and \(\bm{O}^- = \bm{O} - \Delta\bm{O}\cdot \mathbf{1}\), where \(\mathbf{1}\) equals 1 for all elements. Based on the mean value theorem \cite{stewart2015calculus}, there exists an \(\bm{O}'\) between \(\bm{O}^-\) and \(\bm{O}^+\) such that
\begin{equation} \label{eq:hessian}
    \bar{\bm{H}}^{(\bm{O}')}_{i,i} = \frac{\bar{\bm{J}}_i^{(\bm{O}^+)} - \bar{\bm{J}}_i^{(\bm{O}^-)}}{2\cdot \Delta\bm{O}_i},
\end{equation}
where \(\bar{\bm{J}}^{(\bm{O}^+)}\) and \(\bar{\bm{J}}^{(\bm{O}^-)}\) are the Jacobian matrices at \(\bm{O}^+\) and \(\bm{O}^-\), which are computed through backward-propagation. As the perturbation \(\Delta\bm{O}\) is small enough, we approximate \(\bar{\bm{H}}^{(\bm{O})}_{i,i}\) by \(\bar{\bm{H}}^{(\bm{O}')}_{i,i}\). The perturbation Hessian loss is thus formulated as below:
\begin{equation} \label{eq:hessian_instance}
    \mathcal{L}_{\mathrm{PH}}=\sum_i \left(\widehat{\bm{O}}_i - \bm{O}_i\right)^2\cdot \bar{\bm{H}}^{(\bm{O})}_{i,i}.
\end{equation}

It is worth noting that using distinct Hessians for different samples may lead to an unstable training process. To address this issue, we compute the average Hessian across all samples and utilize the mean value to formulate the final reconstruction loss as below:
\begin{equation}\label{eq:aph}
    \bar{\bm{H}}_{i,i}=\frac{1}{N}\sum_{n=1}^{N} \bar{\bm{H}}^{(\bm{O}^{(n)})}_{i,i},
\end{equation}
\begin{equation}\label{eq:l_aph} 
    \mathcal{L}_{\mathrm{APH}} = \sum_i \left(\widehat{\bm{O}}_i - \bm{O}_i\right)^2\cdot \bar{\bm{H}}_{i,i}, 
\end{equation}
where \(\bm{O}^{(n)}\) is the output of the $n-$th sample, and \(N\) is the sample size. 

Ideally, \(\mathcal{L}_{\mathrm{APH}}\) and \(\mathcal{L}_{\mathrm{PH}}\) have the following properties.

\begin{theory}
The expectation of the APH loss is consistent with that of the PH loss, \ie, \(\mathbb{E}\left[\mathcal{L}_{\mathrm{APH}}\right] = \mathbb{E}\left[\mathcal{L}_{\mathrm{PH}}\right] \), under certain independence assumptions.
\end{theory}

\begin{theory}
    When utilizing mini-batch gradient descent, the variance of the gradient of the quantization parameter \(\theta\) w.r.t. the APH loss is smaller than that of the PH loss under certain independence assumptions:
    \begin{equation}
        \mathrm{Var}\left[\frac{\partial \mathcal{L}_{\mathrm{APH}}}{\partial \theta}\right] \leq \mathrm{Var}\left[\frac{\partial \mathcal{L}_{\mathrm{PH}}}{\partial \theta}\right].
    \end{equation}
\end{theory}

We refer to Sec. \ref{sec:proof_1} and  Sec. \ref{sec:proof_2} of the \emph{supplementary material} for detailed proof. Theorems 3.1 and 3.2 imply that the gradient of the APH loss is an unbiased estimation on that of the PH loss, while effectively reducing its variance, under certain independence assumptions. As claimed in \cite{svrg,scsg}, lower gradient variance results in faster convergence and improved training stability. Therefore, our proposed APH loss is expected to outperform the PH loss.

Compared to the Hessian in BRECQ, our method only requires one additional forward and backward pass, while maintaining the same training complexity. As a result, the extra computational overhead is negligible. The key advantages of our method lie in two-fold: 1) APH is deduced directly from the definition, thus eliminating errors introduced by the Fisher Information Matrix; 2) APH is theoretically generalizable to other tasks besides classification, such as object detection and segmentation.

\subsection{MLP Reconstruction} \label{sec:mlp_recon}

As depicted in Sec.~\ref{sec:intro}, quantizing post-GELU activations in ViTs incurs two significant challenges: 1)~the post-GELU activation distribution is highly imbalanced, \ie, concentrating within the narrow interval (-0.17, 0], which leads to approximation errors during quantization \cite{PTQ4ViT}. 2)~the activation range of post-GELU activations varies substantially.

In this section, we propose an MLP Reconstruction method to address the above two issues simultaneously. To deal with the imbalanced distribution, we replace all GELU activation functions in MLP with ReLU. Subsequently, we perform the feature knowledge distillation \cite{kd}, and reconstruct MLP individually. Specifically, for each MLP, we obtain its original input and output using the unlabeled data. By following \cref{sec:aph}, we compute the average perturbation Hessian to determine the output importance. Thereafter, we replace the MLP activation function with ReLU and utilize the Hessian importance to calculate the weighted \(L_2\) distance between the output with ReLU and the original one with GELU, formulated as below:
\begin{equation}\label{eq:o_direct}
    \mathcal{L}_{\mathrm{Direct}} = \left(\bm{O}_{\mathrm{GELU}} - \bm{O}_{\mathrm{Direct}}\right)^2 \odot \bar{\bm{H}},
\end{equation}
where \( \bm{O}_{\mathrm{Direct}} = \mathrm{FC2}(\mathrm{ReLU}(\mathrm{FC1}(\bm{X})))\) is the output of reconstructed MLP with ReLU for input \(\bm{X}\), \(\bm{O}_{\mathrm{GELU}}\) is the output of the original MLP with GELU, and \(\bar{\bm{H}}\) is the average perturbation Hessian.

The reason ReLU can be used as a replacement for GELU lies in the fact that, in deeper Transformers, ReLU may suffer from the dying ReLU problem \cite{dyingrelu}, which is why GELU is typically used during training. However, as described in \cite{reluexpress}, neural networks with ReLU activations also theoretically possess universal approximation capabilities. In this paper, the MLP module are reconstructed individually for each layer, which is of shallow depth, thus avoiding the dying ReLU problem. This enables the network to achieve expressive capability comparable to that by using GELU.

\begin{table*}[t]
\setlength\tabcolsep{4.5pt}
\centering
\caption{Comparison of the top-1 accuracy (\%) on the ImageNet dataset with different quantization bit-widths. Here `Opt.' means whether or not using an optimize-based PTQ method, `PSQ' refers to `Post-Softmax Quantizer', and `PGQ' refers to `Post-GELU Quantizer'. `*' indicates that the results are reproduced by using the official code. `TUQ', `MPQ', `GUQ', `SULQ', and `TanQ' are the abbreviations of `Twin-Uniform Quantizer' in PTQ4ViT, `Matthew-effect Preserving Quantizer' in APQ-ViT, `Groupwise Uniform Quantizer' in IGQ-ViT, `Shift-Uniform-Log2 Quantizer' in I\&S-ViT, and `Tangent Quantizer' in DopQ-ViT, respectively.}
\begin{tabular}{cccccccccccc}
   \toprule
   {\textbf{Method}} & \textbf{Opt.} & \textbf{PSQ} & \textbf{PGQ} & \textbf{W/A} & \textbf{ViT-S} & \textbf{ViT-B} & \textbf{DeiT-T} & \textbf{DeiT-S} & \textbf{DeiT-B} & \textbf{Swin-S} & \textbf{Swin-B}\\

   \midrule
   Full-Prec. & - & - & - & 32/32 & 81.39 & 84.54 & 72.21 & 79.85 & 81.80 & 83.23 & 85.27\\
   \midrule
   PTQ4ViT \cite{PTQ4ViT} & $\times$ & TUQ & TUQ & 3/3 & 0.10 & 0.10 & 3.50 & 0.10 & 31.06 & 28.69 & 20.13 \\
   RepQ-ViT \cite{RepQViT} & $\times$ & $\log \sqrt{2}$ & Uniform & 3/3 & 0.10 & 0.10 & 0.10 & 0.10 & 0.10 & 0.10 & 0.10 \\
   AdaLog \cite{adalog} & $\times$ & AdaLog & AdaLog & 3/3 & 13.88 & 37.91 & 31.56 & 24.47 & 57.47 & 64.41 & 69.75 \\
   I\&S-ViT \cite{isvit} & \checkmark & SULQ & Uniform & 3/3 & 45.16 & 63.77 & 41.52 & 55.78 & 73.30 & 74.20 & 69.30 \\
   DopQ-ViT \cite{dopq} & \checkmark & TanQ & Uniform & 3/3 & 54.72 & 65.76 & 44.71 & 59.26 & 74.91 & 74.77 & 69.63 \\
   QDrop* \cite{qdrop} & \checkmark & Uniform & Uniform & 3/3 & 38.31 & 73.79 & 46.69 & 52.55 & 74.32 & 74.11 & 75.28 \\
\rowcolor{lightgray!45} \textbf{APHQ-ViT(Ours)} & \checkmark & Uniform & Uniform & 3/3 & \textbf{63.17} & \textbf{76.31} & \textbf{55.42} & \textbf{68.76} & \textbf{76.31} & \textbf{76.10} & \textbf{78.14} \\
   \midrule
   PTQ4ViT \cite{PTQ4ViT} & $\times$ & TUQ & TUQ & 4/4 & 42.57 & 30.69 & 36.96 & 34.08 & 64.39 & 76.09 & 74.02  \\
   APQ-ViT \cite{APQViT} & $\times$ & MPQ & Uniform & 4/4 & 47.95 & 41.41 & 47.94 & 43.55 & 67.48 & 77.15 & 76.48 \\
   RepQ-ViT \cite{RepQViT} & $\times$ & $\log \sqrt{2}$ & Uniform & 4/4 & 65.05 & 68.48 & 57.43 & 69.03 & 75.61 & 79.45 & 78.32 \\
   ERQ \cite{erq} & $\times$ & $\log \sqrt{2}$ & Uniform & 4/4 & 68.91 & 76.63 & 60.29 & 72.56 & 78.23 & 80.74 & 82.44 \\
   IGQ-ViT \cite{igqvit} & $\times$ & GUQ & GUQ & 4/4 & 73.61 & 79.32 & 62.45 & 74.66 & 79.23 & 80.98 & 83.14 \\
   AdaLog \cite{adalog} & $\times$ & AdaLog & AdaLog & 4/4 & 72.75 & 79.68 & 63.52 & 72.06 & 78.03 & 80.77 & 82.47 \\ 
   I\&S-ViT \cite{isvit} & \checkmark & SULQ & Uniform & 4/4 & 74.87 & 80.07 & 65.21 & 75.81 & 79.97 & 81.17 & 82.60 \\
   DopQ-ViT \cite{dopq} & \checkmark & TanQ & Uniform & 4/4 & 75.69 & 80.95 & 65.54 & 75.84 & 80.13 & 81.71 & 83.34 \\
   QDrop* \cite{qdrop} & \checkmark & Uniform & Uniform & 4/4 & 67.62 & 82.02 & 64.95 & 74.73 & 79.64 & 81.03 & 82.79 \\
   OASQ \cite{oasq} & \checkmark & Unifrom & Uniform & 4/4 & 72.88 & 76.59 & 66.31 & 76.00 & 78.83 & 81.02 & 82.46 \\
\rowcolor{lightgray!45} \textbf{APHQ-ViT(Ours)} & \checkmark & Uniform & Uniform & 4/4 & \textbf{76.07} & \textbf{82.41} & \textbf{66.66} & \textbf{76.40} & \textbf{80.21} & \textbf{81.81}& \textbf{83.42} \\
   \bottomrule
\end{tabular}
\label{tab:classification}
\end{table*}

To address the activation range issue, we design an alternative clamp loss to constrain the range effectively. Specifically, we compute the \(p\)-th percentile of all positive values and restrict the activations within this \(p\)-th percentile. The clipped output is formulated as:
\begin{align}
\begin{split} \label{eq:mlprecon}
    \bm{A}_{\mathrm{FC2}} &= \mathrm{ReLU}(\mathrm{FC1}(\bm{X})), \\
    \bm{O}_{\mathrm{clamp}} &= \mathrm{FC2}(\mathrm{clamp}(\bm{A}_{\mathrm{FC2}},\ \mathrm{Quantile}_{p}(\bm{A}_{\mathrm{FC2}}))). \\
\end{split}
\end{align}
Accordingly, the clipped reconstruction loss is written as:
\begin{equation}
    \mathcal{L}_{\mathrm{Clamp}} = \left(\bm{O}_{\mathrm{GELU}} - \bm{O}_{\mathrm{clamp}}\right)^2 \odot \bm{H}.
\end{equation}
The MLP Reconstruction loss is finally formulated as:
\begin{equation} \label{eq:l_distill}
    \mathcal{L}_{\mathrm{Distill}} = \mathcal{L}_{\mathrm{Direct}} + \alpha \cdot \mathcal{L}_{\mathrm{Clamp}},
\end{equation}
where \(\alpha\) is a trade-off hyperparameter fixed as \(\alpha=2\). It is important to note that \(\mathcal{L}_{\mathrm{Direct}}\) cannot be omitted. By solely using \(\mathcal{L}_{\mathrm{Clamp}}\) leads to vanishing gradients for hard-clipped activations. In regions where activations are hard-clipped, the gradients tend to be zero, hindering the effective update of the MLP parameters. By incorporating \(\mathcal{L}_{\mathrm{Direct}}\), which leverages unclamped activations, the gradient vanishing issue is mitigated, thus facilitating effective learning.
 
\section{Experimental Results and Analysis}

\subsection{Experimental Setup} \label{sec:detail}
\noindent \textbf{Datasets and Models.}~For the classification task, we evaluate our method on ImageNet~\cite{imagenet} with representative Vision Transformer architectures, including ViT~\cite{vit}, DeiT~\cite{deit} and Swin~\cite{swin}. For object detection and instance segmentation, we evaluate on COCO~\cite{coco} by utilizing the Mask R-CNN~\cite{maskrcnn} and Cascade Mask R-CNN~\cite{cascadercnn} frameworks based on the Swin backbones. 

\noindent \textbf{Implementation Details.}~All pretrained full-precision Vision Transformers are obtained from the timm library\footnote{https://github.com/huggingface/pytorch-image-models}. The pretrained detection and segmentation models are obtained from MMDetection \cite{mmdetection}. Following existing works \cite{BRECQ,qdrop,oasq,isvit}, we randomly select 1024 unlabeled images from ImageNet and 256 unlabeled images from COCO as the calibration sets for classification and object detection, respectively. We adopt channel-wise uniform quantizers for weight quantization and layer-wise uniform quantizers for activation quantization, including the attention map. We follow the hyper-parameter settings as used in QDrop \cite{qdrop} by setting the batch size, learning rate for activation quantization, learning rate for tuning weight, the maximal iteration number in both MLP Reconstruction and QDrop reconstruction as 32, 4e-5, 1e-3 and 20000, respectively. In addition, we set the percentile \(p=0.99\) in \cref{eq:mlprecon}.

\subsection{Quantization Results on ImageNet}

We first compare our method to the state-of-the-art approaches for post-training quantization of ViTs on ImageNet: 1) the calibration-only methods including PTQ4ViT~\cite{PTQ4ViT}, APQ-ViT \cite{APQViT}, RepQ-ViT~\cite{RepQViT}, ERQ \cite{erq}, IGQ-ViT \cite{igqvit} and AdaLog \cite{adalog}; and 2) the reconstruction-based methods including DopQ-ViT~\cite{dopq}, QDrop \cite{qdrop} and OASQ~\cite{oasq}. 

\begin{table*}[t]
\setlength\tabcolsep{7pt}
\centering
\caption{Quantization results (\%) on COCO for the object detection and instance segmentation tasks. Here, `Baseline' refers to the results by using only uniform quantizers for calibration. * and $\dagger$ indicate that the results are re-produced by using the official code.}
\begin{tabular}{cccccccccccc}
   \toprule
   \multirow{3} * {\textbf{Method}} & \multirow{3} * {\textbf{Opt.}} & \multirow{3} * {\textbf{PSQ}} & \multirow{3} * {\textbf{W/A}} & \multicolumn{4}{c}{\textbf{Mask R-CNN}} & \multicolumn{4}{c}{\textbf{Cascade Mask R-CNN}} \\
   \cmidrule(lr){5-8}\cmidrule(lr){9-12}
   ~ & ~ & ~ & ~ & \multicolumn{2}{c}{\textbf{Swin-T}} & \multicolumn{2}{c}{\textbf{Swin-S}} & \multicolumn{2}{c}{\textbf{Swin-T}} & \multicolumn{2}{c}{\textbf{Swin-S}} \\
   ~ & ~ & ~ & ~ & AP\textsuperscript{b} & AP\textsuperscript{m} & AP\textsuperscript{b} & AP\textsuperscript{m} & AP\textsuperscript{b} & AP\textsuperscript{m} & AP\textsuperscript{b} & AP\textsuperscript{m} \\
   \midrule
   Full-Precision & - & - & 32/32  & 46.0 & 41.6 & 48.5 & 43.3 & 50.4 & 43.7 & 51.9 & 45.0 \\
   \midrule
   Baseline* & $\times$ & Uniform & 4/4 & 34.6 & 34.2 & 40.8 & 38.6 & 45.9 & 40.2 & 47.9 & 41.6 \\
   RepQ-ViT \cite{RepQViT} & $\times$ & $\log\sqrt{2}$ & 4/4 & 36.1 & 36.0 & $44.2_{42.7}\dagger$ & 40.2 & 47.0 & 41.1 & 49.3 & 43.1 \\
   ERQ \cite{erq} & $\times$ & $\log\sqrt{2}$ & 4/4 & 36.8 & 36.6 & 43.4 & 40.7 & 47.9 & 42.1 & 50.0 & 43.6 \\
   I\&S-ViT \cite{isvit} & \checkmark & SULQ & 4/4 & 37.5 & 36.6 & 43.4 & 40.3 & 48.2 & 42.0 & \textbf{50.3} & 43.6 \\
   DopQ-ViT \cite{dopq} & \checkmark & TanQ & 4/4 & 37.5 & 36.5 & 43.5 & 40.4 & 48.2 & 42.1 & \textbf{50.3} & \textbf{43.7} \\
   QDrop* \cite{qdrop} & \checkmark & Uniform & 4/4 & 36.2 & 35.4 & 41.6 & 39.2 & 47.0 & 41.3 & 49.0 & 42.5 \\
    \rowcolor{lightgray!45}\textbf{APHQ-ViT (Ours)} & \checkmark & Uniform & 4/4 & \textbf{38.9} & \textbf{38.1} & \textbf{44.1} & \textbf{41.0} & \textbf{48.9} & \textbf{42.7} & \textbf{50.3} & \textbf{43.7} \\
   \bottomrule
\end{tabular}
\label{tab:detection}
\end{table*}

As summarized in Table~\ref{tab:classification}, for 4-bit quantization, some of the compared methods suffer a remarkable degradation in accuracy due to severe quantization loss of weights and activations. However, the performance of the proposed APHQ-ViT remains competitive compared to the full-precision models and consistently outperforms existing methods. As for 3-bit quantization, calibration-only methods yield an extremely low performance (\emph{e.g.}~0.1\%) in most scenarios. Reconstruction-based methods like DopQ-ViT and I\&S-ViT also suffer significant accuracy loss on models that are challenging to quantize (\emph{e.g.} ViT-S and DeiT-T). By contrast, APHQ-ViT maintains more stable accuracy when reducing the precision from 32 bits to 3 bits. It surpasses the second-best method, DopQ-ViT, by \(10.71\%\) when using the DeiT-T backbone and achieves an average improvement of $7.21\%$. 

\subsection{Quantization Results on COCO}

We further evaluate our method on COCO for object detection and instance segmentation. As shown in Table~\ref{tab:detection}, the baseline method, which employs only uniform quantizers and QDrop, achieves lower accuracy compared to other calibration-only and reconstruction-based methods that utilize specific quantizers. By employing the APH loss and MLP Reconstruction, our method achieves results on par with or superior to those using specific quantizers.

\subsection{Ablation Studies}

\begin{table}[t]
\setlength\tabcolsep{4pt}
\centering
\caption{Ablation results w.r.t the top-1 accuracy (\%) of the proposed main components on ImageNet with the W3/A3 setting.}
\begin{tabular}{cccccc}
    \toprule
    \textbf{Method} & \textbf{ViT-S} & \textbf{ViT-B} & \textbf{DeiT-T} & \textbf{DeiT-S} & \textbf{Swin-S} \\
    \cmidrule(lr){1-6}
    Full-Prec. & 81.39 & 84.54 & 72.21 & 79.85 & 81.80 \\
    \midrule
     QDrop      & 38.31 & 73.79 & 46.69 & 52.55 & 74.11 \\
     +APH          & 59.11 & 76.05 & 53.82 & 67.40 & 75.44 \\
     +APH +MR & \textbf{63.17} & \textbf{76.31} & \textbf{55.42} & \textbf{68.76} & \textbf{76.10} \\
   \bottomrule
\end{tabular}
\label{tab:ablation_main}
\end{table}

\textbf{Effect of the Main Components.} We first evaluate the effectiveness of the proposed Average Perturbation Hessian (APH) loss and the MLP Reconstruction (MR) method. As displayed in Table~\ref{tab:ablation_main}, applying the APH loss on QDrop reconstruction significantly promotes the top-1 accuracy across distinct Vision Transformer architectures. Specifically, the accuracy is improved by 20.80\%, 14.85\%, and 7.13\% when using ViT-S, DeiT-S, and DeiT-T on W3/A3, respectively. MLP Reconstruction consistently boosts the accuracy when combined with the APH loss.

\textbf{Average Perturbation Hessian.} To validate the effectiveness of the proposed APH loss, we compare it with the alternative representative quantization loss, including the MSE loss \cite{qdrop} and the BRECQ based Hessian (BH) loss. We further compare it with the original Perturbation Hessian (PH) without averaging. As shown in Table~\ref{tab:ab_hessian}, the PH loss outperforms other quantization losses in most ViT architectures, and the APH loss further improves the accuracy.

\textbf{MLP Reconstruction.} We separately reconstruct the MLP module, \emph{i.e.}, performing MLP Reconstruction one by one without utilizing QDrop reconstruction. As summarized in Table~\ref{tab:ab_mr}, except for a performance drop of over 1\% on DeiT-T, the accuracy loss on other models is less than 0.5\%. On ViT-B, the accuracy even surpasses that of the full-precision model by adopting the MR method.

We provide more ablation results in Sec.~\ref{sec:more_res} of the \emph{supplementary material}.

\begin{table}[t]
\setlength\tabcolsep{4.8pt}
\centering
\caption{Ablation results w.r.t the top-1 accuracy (\%) of the proposed  Perturbation Hessian, compared to other losses on ImageNet with the W3/A3 setting. ``BH", ``PH" and ``APH" denote ``BRECQ-based Hessian", ``Perturbation Hessian"  and ``Average Perturbation Hessian", respectively.}
\begin{tabular}{cccccc}
    \toprule
    \textbf{Method} & \textbf{ViT-S} & \textbf{ViT-B} & \textbf{DeiT-T} & \textbf{DeiT-S} & \textbf{Swin-S} \\
    \midrule
    Full-Prec. & 81.39 & 84.54 & 72.21 & 79.85 & 83.23 \\
    \midrule
     MSE \cite{qdrop} & 38.31 & 73.79 & 46.69 & 52.55 & 74.11 \\
     BH \cite{BRECQ}  & 54.33 & 66.62 & 49.27 & 63.72 & 75.20 \\
     PH & 55.14 & 72.80 & 52.25 & 66.12 &  75.40 \\
     \textbf{APH} & \textbf{59.11} & \textbf{76.05} & \textbf{53.82} & \textbf{67.40} & \textbf{75.44} \\
   \bottomrule
\end{tabular}
\label{tab:ab_hessian}
\end{table}

\begin{table}[t]
\setlength\tabcolsep{3.8pt}
\centering
\caption{Ablation results w.r.t the top-1 accuracy (\%) of the proposed MLP Reconstruction method on ImageNet.}
\begin{tabular}{cccccc}
    \toprule
    \textbf{Method} & \textbf{ViT-S} & \textbf{ViT-B} & \textbf{DeiT-T} & \textbf{DeiT-S} & \textbf{Swin-S} \\
    \cmidrule(lr){1-6}
    Full-Prec. & 81.39 & 84.54 & 72.21 & 79.85 & 83.23 \\
    \midrule
    MLP Recon.      & 80.90	& 84.84	& 71.07	& 79.38 & 83.12 \\
   \bottomrule
\end{tabular}
\label{tab:ab_mr}
\end{table}

\subsection{Analysis of Inference Efficiency on MR}

\begin{table}[t]
    \centering
    \caption{Comparison of latency and throughput of ViTs under W8A8 quantization to full-precision models. ``AF" indicates the adopted activation function. ``Lat." refers to the model latency (in milliseconds). ``TP" stands for the throughput (in images per second). ``SR" is the speedup rate.}
    \begin{tabular}{cccccc}
        \toprule
        \multirow{2}{*}{\begin{minipage}{0.95cm}\centering \textbf{Model} \end{minipage} } & 
        \multirow{2}{*}{\begin{minipage}{0.6cm}\centering \textbf{AF} \end{minipage} } & 
        \multirow{2}{*}{\begin{minipage}{0.6cm}\centering \textbf{Bits} \end{minipage} } & 
        \multirow{2}{*}{\begin{minipage}{0.6cm}\centering \textbf{Lat.} \end{minipage} } & 
        \multirow{2}{*}{\begin{minipage}{0.95cm}\centering \textbf{TP}\end{minipage} } & 
        \multirow{2}{*}{\begin{minipage}{0.7cm}\centering \textbf{SR}\end{minipage} } \\
        ~ \\
        \midrule
        \multirow{3}{*}{\begin{minipage}{1.1cm}\centering \textbf{DeiT-T} \end{minipage} } & GELU & 32 & 30.93 & 32.08 & $\times 1$ \\
        ~ & GELU & 8 & 22.34 & 44.76 & $\times$ 1.40 \\
        ~ & ReLU & 8 & \textbf{20.66} & \textbf{48.40} & $\times$ \textbf{1.51} \\
        \midrule
        \multirow{3}{*}{\begin{minipage}{1.1cm}\centering \textbf{DeiT-S} \end{minipage} } & GELU & 32 & 100.03 & 9.97 & $\times 1$ \\
        ~ & GELU & 8 & 63.89 & 15.65 & $\times$ 1.57 \\
        ~ & ReLU & 8 & \textbf{58.40} & \textbf{17.12} & $\times$ \textbf{1.72} \\
        \midrule
        \multirow{3}{*}{\begin{minipage}{1.1cm}\centering \textbf{DeiT-B} \end{minipage} } & GELU & 32 & 346.93 & 2.88 & $\times 1$ \\
        ~ & GELU & 8 & 217.96 & 4.59 & $\times$ 1.59 \\
        ~ & ReLU & 8 & \textbf{198.80} & \textbf{5.03} & $\times$ \textbf{1.75} \\
        \midrule
        \multirow{3}{*}{\begin{minipage}{1.1cm}\centering \textbf{Swin-S} \end{minipage} } & GELU & 32 & 255.56 & 3.90 & $\times 1$ \\
        ~ & GELU & 8 & 180.42 & 5.54 & $\times$ 1.42 \\
        ~ & ReLU & 8 & \textbf{171.88} & \textbf{5.82} & $\times$ \textbf{1.49} \\
        \midrule
        \multirow{3}{*}{\begin{minipage}{1.1cm}\centering \textbf{Swin-B} \end{minipage} } & GELU & 32 & 411.07 & 2.43 & $\times 1$ \\
        ~ & GELU & 8 & 282.28 & 3.54 & $\times$ 1.44 \\
        ~ & ReLU & 8 & \textbf{264.38} & \textbf{3.78} & $\times$ \textbf{1.54} \\
        \bottomrule
    \end{tabular}
    \label{tab:latency}
\end{table}

MLP Reconstruction replaces the GELU activation function with ReLU. Unlike GELU, which incurs additional computational overhead, ReLU can be folded into the preceding linear layer. As a consequence, the proposed MR method not only promotes quantization accuracy but also accelerates inference. Since quantization below 8 bits typically requires specialized hardware \cite{erq,packqvit,BRECQ}, we benchmark the quantized model at W8A8 on an Intel i5-12400F CPU. As shown in Table \ref{tab:latency}, 8-bit quantization generally achieves a 1.4 to 1.6 times speedup. By replacing GELU with ReLU via MR, we further improve inference efficiency.

\subsection{Discussion on Training Efficiency}

\begin{table}[t]
    \centering
    \setlength\tabcolsep{4.5pt}
    \caption{Comparison of the training time cost and accuracy (\%) under W3/A3 by using distinct quantization methods on a single Nvidia RTX 4090 GPU.}
    \begin{tabular}{cccccc}
        \toprule
        \multirow{2}{*}{\begin{minipage}{1.1cm}\centering \textbf{Model} \end{minipage} } & 
        \multirow{2}{*}{\begin{minipage}{1.1cm}\centering \textbf{Method} \end{minipage} } & 
        \multirow{2}{*}{\begin{minipage}{0.6cm}\centering \textbf{PTQ} \end{minipage} } & 
        \multirow{2}{*}{\begin{minipage}{0.65cm}\centering \textbf{Data} \\ \textbf{Size} \end{minipage} }  & 
        \multirow{2}{*}{\begin{minipage}{0.65cm}\centering \textbf{Time} \\ \textbf{Cost} \end{minipage} }  & 
        \multirow{2}{*}{\begin{minipage}{0.6cm}\centering \textbf{Acc.} \end{minipage} }\\
        ~ \\
        \midrule
        \multirow{3}{*}{\begin{minipage}{1.1cm}\centering \textbf{DeiT-S} \end{minipage} } & LSQ \cite{stepsize} & $\times$ & 1280 K & $ \sim $170 h & 77.3 \\
        ~ & QDrop \cite{qdrop} & \checkmark & 1024 & 47 min & 52.6 \\
        ~ & APHQ-ViT & \checkmark & 1024 & 62 min & 68.8 \\
        \midrule
        \multirow{3}{*}{\begin{minipage}{1.1cm}\centering \textbf{Swin-S} \end{minipage} } & LSQ \cite{stepsize} & $\times$ & 1280 K & $ \sim $450 h & 80.6 \\
        ~ & QDrop \cite{qdrop} & \checkmark & 1024 & 130 min & 74.1 \\
        ~ & APHQ-ViT & \checkmark & 1024 & 170 min & 76.1 \\
        \bottomrule
    \end{tabular}
    \label{tab:traincost}
\end{table}

The MLP Reconstruction method in APHQ-ViT introduces additional training overhead. However, the extra training cost is acceptable. As shown in Table \ref{tab:traincost}, our method incurs less training overhead, compared to QAT methods such as LSQ. Furthermore, our approach requires only 1024 unlabeled images as a calibration set, eliminating fine-tuning on the entire dataset, as is typically required by QAT methods.

\section{Conclusion}
\label{sec:conclusion}

In this paper, we propose a novel post-training quantization approach dubbed APHQ-ViT for Vision Transformers. We first demonstrate that the current Hessian guided loss adopts an inaccurate estimated Hessian matrix, and present an improved Average Perturbation Hessian (APH) loss. Based on APH, we develop an MLP Reconstruction method that simultaneously replaces the GELU activation function with ReLU and significantly reduces the activation range. Extensive experimental results show the effectiveness of our approach across various Vision Transformer architectures and vision tasks, including image classification, object detection, and instance segmentation. Notably, compared to the state-of-the-art methods, APHQ-ViT achieves an average improvement of 7.21\% on ImageNet with 3-bit quantization using only uniform quantizers. 

\section*{Acknowledgments}
This work was partly supported by the Beijing Municipal Science and Technology Project (No. Z231100010323002), the National Natural Science Foundation of China (Nos. 62202034,62176012,62022011,62306025), the Beijing Natural Science Foundation (No. 4242044), the Aeronautical Science Foundation of China (No. 2023Z071051002), CCF-Baidu Open Fund, the Research Program of State Key Laboratory of Virtual Reality Technology and Systems, and the Fundamental Research Funds for the Central Universities.  

{
    \small
    \bibliographystyle{ieeenat_fullname}
    \bibliography{main}
}

\appendix \setcounter{page}{1}
\maketitlesupplementary

\renewcommand{\thetable}{\Alph{table}}
\setcounter{table}{0}

\renewcommand{\thefigure}{\Alph{figure}}
\setcounter{figure}{0}

In this document, we provide detailed proofs on Theorem 3.1 and Theorem 3.2 in the main body in Sec.~\ref{sec:main_proof}, and provide more ablation studies and visualization results in Sec.~\ref{sec:more_res} and Sec.~\ref{sec:visualize}, respectively.

\section{Main Proofs} \label{sec:main_proof}
\subsection{Proof of Theorem 3.1} \label{sec:proof_1}

\begin{proof}
In regards of the perturbation Hessian \(\mathcal{L}_{\mathrm{PH}}\), we can deduce the following equation:
\begin{align}
\begin{split}
\mathbb{E}\left[\mathcal{L}_{\mathrm{PH}}\right]
    &=  \mathbb{E}_{(\bm{O}, \theta_q)}\left[ \sum_{i}\left( \widehat{\bm{O}}_{i}^{(k,\theta_q)}-\bm{O}_{i}^{(k)} \right)^2 \cdot \bar{\bm{H}}_{i,i}^{(\bm{O}^{(k)})}\right] \\
    &= \sum_{i} \mathbb{E}_{(\bm{O}, \theta_q)}\left[ \left( \widehat{\bm{O}}_{i}^{(k,\theta_q)}-\bm{O}_{i}^{(k)} \right)^2\cdot \bar{\bm{H}}_{i,i}^{(\bm{O}^{(k)})}\right].  
\end{split}
\end{align}
where \(\theta_q\) denotes the quantization parameter. Since the Hessian matrix is computed by adding fixed perturbations to the output, it is an inherent attribute of the networks. Thus, we assume that the Hessian matrix is independent of \(\widehat{\bm{O}}_{i}^{(k, \theta_q)}-\bm{O}_{i}^{(k)}\), and the following equation holds:
\begin{align}\label{eq:PH}
\begin{split}
\mathbb{E}\left[\mathcal{L}_{\mathrm{PH}}\right]
    &= \sum_{i} \mathbb{E}\left[ \left( \widehat{\bm{O}}_{i}^{(k)}-\bm{O}_{i}^{(k)} \right)^2\right] \cdot \mathbb{E}\left[\bar{\bm{H}}_{i,i}^{(\bm{O}^{(k)})}\right]. \\
\end{split}
\end{align}
As for the average perturbation Hessian, the following equations hold:
\begin{align}\label{eq:APH}
\begin{split}
\mathbb{E}\left[\mathcal{L}_{\mathrm{APH}}\right]
    &= \mathbb{E}_{(\bm{O}, \theta_q)}\left[ \sum_{i} \left( \widehat{\bm{O}}_{i}^{(k, \theta_q)}-\bm{O}_{i}^{(k)} \right)^2 \cdot \bar{\bm{H}}_{i,i}\right]\\
    &= \sum_{i}\mathbb{E}\left[ \left( \widehat{\bm{O}}_{i}^{(k, \theta_q)}-\bm{O}_{i}^{(k)} \right)^2\right] \cdot \mathbb{E}\left[\bar{\bm{H}}_{i,i}\right].
\end{split}
\end{align}
Based on Eqs.~\eqref{eq:PH}-\eqref{eq:APH} and \(\mathbb{E}\left[\bar{\bm{H}}_{i,i}^{(\bm{O}^{(k)})}\right]=\mathbb{E}\left[\bar{\bm{H}}_{i,i}\right]\), we can deduce that \(\mathbb{E}\left[\mathcal{L}_{\mathrm{PH}}\right] = \mathbb{E}\left[\mathcal{L}_{\mathrm{APH}}\right] \).
\end{proof}

\subsection{Proof of Theorem 3.2} \label{sec:proof_2}

\begin{proof}
We firstly denote the gradient of the perturbation Hessian (PH) loss w.r.t. the quantization parameter \(\theta_q\)  during the mini-batch gradient descent as below:
\begin{equation}\label{eq:g_theta}
    g(\theta_q) =  \frac{2}{\lvert B\rvert} \sum_{k,i} \bar{\bm{H}}_{i,i}^{(\bm{O}^{(k)})}\left( \widehat{\bm{O}}_{i}^{(k)}-\bm{O}_{i}^{(k)} \right) \frac{\partial \left( \widehat{\bm{O}}_{i}^{(k)}-\bm{O}_{i}^{(k)} \right)}{\partial \theta_q},
\end{equation}
where \(\lvert B\rvert\) is the batch size. We further define the random variable \(X_{i}^{(k)}\):
\begin{equation}
    X_{i}^{(k)} = 2\left( \widehat{\bm{O}}_{i}^{(k)}-\bm{O}_{i}^{(k)} \right) \frac{\partial \left( \widehat{\bm{O}}_{i}^{(k)}-\bm{O}_{i}^{(k)} \right)}{\partial \theta_q}.
\end{equation}
Accordingly, Eq.~\eqref{eq:g_theta} can be rewritten as
\begin{equation}
    g(\theta_q) = \frac{1}{\lvert B\rvert} \sum_{i}\sum_{k} X_{i}^{(k)}\cdot \bar{\bm{H}}_{i,i}^{(\bm{O}^{(k)})}.
\end{equation}

Similarly, as \(\bar{\bm{H}}_{i,i}\approx \mathbb{E}[\bar{\bm{H}}^{\bm{O}^{(k)}}_{i,i}]\) when the sample size $N$ becomes large enough. We denote the gradient of the average perturbation Hessian (APH) loss w.r.t. the parameter \(\theta_q\) as below:
\begin{align}
\begin{split}
    \hat{g}(\theta_q) &= \frac{1}{\lvert B\rvert} \sum_{i}\sum_{k} X_{i}^{(k)}\cdot \bar{\bm{H}}_{i,i} \\
    &\approx \frac{1}{\lvert B\rvert} \sum_{i} \left(\sum_{k} X_{i}^{(k)}\cdot \mathbb{E}[\bar{\bm{H}}^{\bm{O}^{(k)}}_{i,i}] \right).
\end{split}
\end{align}

We assume that all the output elements are independent across different samples and channels. Using the variance formula for the product of random variables, the gradient variance of the original PH loss is formulated as below:
\begin{align}
\begin{split}
    \mathrm{Var}\left[g(\theta_q)\right] &= \frac{1}{\lvert B\rvert^2} \sum_{i} \left(\sum_{k} \mathrm{Var} \left[X_{i}^{(k)}\cdot\bar{\bm{H}}_{i,i}^{(\bm{O}^{(k)})}\right] \right) \\
    &= \frac{1}{\lvert B\rvert^2} \sum_{i} \left( \sum_{k} \mathbb{E}[\bar{\bm{H}}^{\bm{O}^{(k)}}_{i,i}]^2\mathrm{Var}[X^{(k)}_{i}] + R \right),
\end{split}
\end{align}
where
\begin{equation}
    R = \mathrm{Var}[X^{(k)}_{i}]\mathrm{Var}[\bar{\bm{H}}^{\bm{O}^{(k)}}_{i,i}] + \mathbb{E}[X^{(k)}_{i}]^2\mathrm{Var}[\bar{\bm{H}}^{\bm{O}^{(k)}}_{i,i}]
\end{equation}
The gradient variance of the APH is:
\begin{align}
\begin{split}
    \mathrm{Var}\left[\hat{g}(\theta_q)\right] &= \frac{1}{\lvert B\rvert^2} \sum_{i} \left(\sum_{k} \mathrm{Var} \left[X^{(k)}_{i}\cdot \mathbb{E}[\bar{\bm{H}}^{\bm{O}(k)}_{i,i}]\right] \right) \\
    &= \frac{1}{\lvert B\rvert^2} \sum_{i} \left( \sum_{k} \mathbb{E}[\bar{\bm{H}}^{\bm{O}(k)}_{i,i}]^2\mathrm{Var}[X^{(k)}_{i}] \right)
\end{split}
\end{align}
As \(R\ge 0\), we can deduce that \(\mathrm{Var}\left[g(\theta_q)\right] \ge \mathrm{Var}\left[g'(\theta_q)\right]\).
\end{proof}

\section{More Ablation Results} \label{sec:more_res}

In this document, we provide more ablation results for DeiT-B and Swin-B as complements to Tables \ref{tab:ablation_main}-\ref{tab:ab_mr} in the main body. The results are summarized in Table~\ref{tab:ablation_main_2}, Table~\ref{tab:ab_hessian_2} and Table~\ref{tab:ab_mr_2}. As displayed, the APH loss can significantly promotes the accuracy, and outperform the alternative losses. The proposed MR method also effectively reconstructs the pretrained model by replacing the GELU activation function with ReLU, without significantly sacrificing the accuracy.

\begin{table}[t]
\setlength\tabcolsep{4pt}
\centering
\caption{Ablation results w.r.t the top-1 accuracy (\%) of the proposed main components on ImageNet with the W3/A3 setting.}
\begin{tabular}{ccc}
    \toprule
    \textbf{Method} & \textbf{DeiT-B} & \textbf{Swin-B} \\
    \cmidrule(lr){1-3}
    Full-Prec. & 84.54 & 85.27 \\
    \midrule
     baseline      & 74.32 & 75.28 \\
     +APH          & 75.62 & 77.16 \\
     +APH +MR & \textbf{76.31} & \textbf{78.14} \\
   \bottomrule
\end{tabular}
\label{tab:ablation_main_2}
\end{table}

\begin{table}[t]
\setlength\tabcolsep{5pt}
\centering
\caption{Ablation results w.r.t the top-1 accuracy (\%) of the proposed APH loss, compared to alternative losses on ImageNet with the W3/A3 setting.}
\begin{tabular}{ccc}
    \toprule
    \textbf{Method} & \textbf{DeiT-B} & \textbf{Swin-B} \\
    \midrule
    Full-Prec. & 81.80 & 85.27 \\
    \midrule
     MSE & 74.32 & 75.28 \\
     BH  & 72.90 & 76.63 \\
     PH & 75.03 & 76.89 \\
     \textbf{APH} & \textbf{75.62} & \textbf{77.16} \\
   \bottomrule
\end{tabular}
\label{tab:ab_hessian_2}
\end{table}

\begin{table}[t]
\setlength\tabcolsep{4pt}
\centering
\caption{Ablation results w.r.t the top-1 accuracy (\%) of the proposed MLP Reconstruction (MR) method on ImageNet with the W3/A3 setting.}
\begin{tabular}{ccc}
    \toprule
    \textbf{Method} & \textbf{DeiT-B} & \textbf{Swin-B} \\
    \cmidrule(lr){1-3}
    Full-Prec. & 81.80 & 85.27 \\
    \midrule
    \textbf{MR}      & 81.43	& 84.97 \\
   \bottomrule
\end{tabular}
\label{tab:ab_mr_2}
\end{table}

\section{Visualization Results} \label{sec:visualize}

\subsection{Loss Curve of APH}

\begin{figure}[t]
    \centering
    \includegraphics[width=1\linewidth]{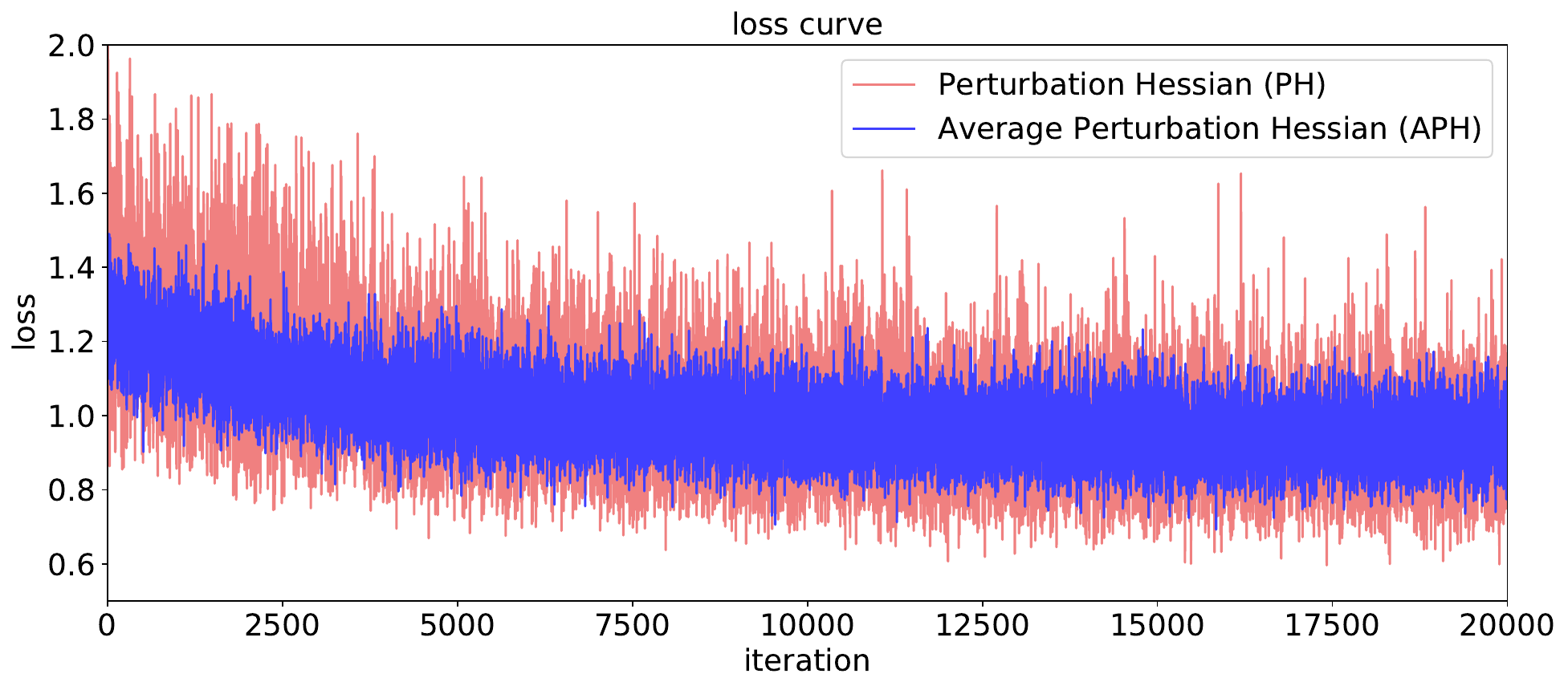}
    \caption{The loss curve of \textit{ViT-Small-blocks.6} on W3/A3.}
    \label{fig:loss}
\end{figure}

\cref{fig:loss} shows the loss curves of the perturbation Hessian (PH) loss and the average perturbation Hessian (APH) loss for a certain block. As illustrated, the APH loss generally exhibits smaller fluctuations than the PH loss, resulting in more stable training.

\subsection{APH Importance} \label{sec:importance}

\begin{figure}[t]
    \centering
    \begin{subfigure}{.50\columnwidth}
        \centering
        \includegraphics[width=1\linewidth]{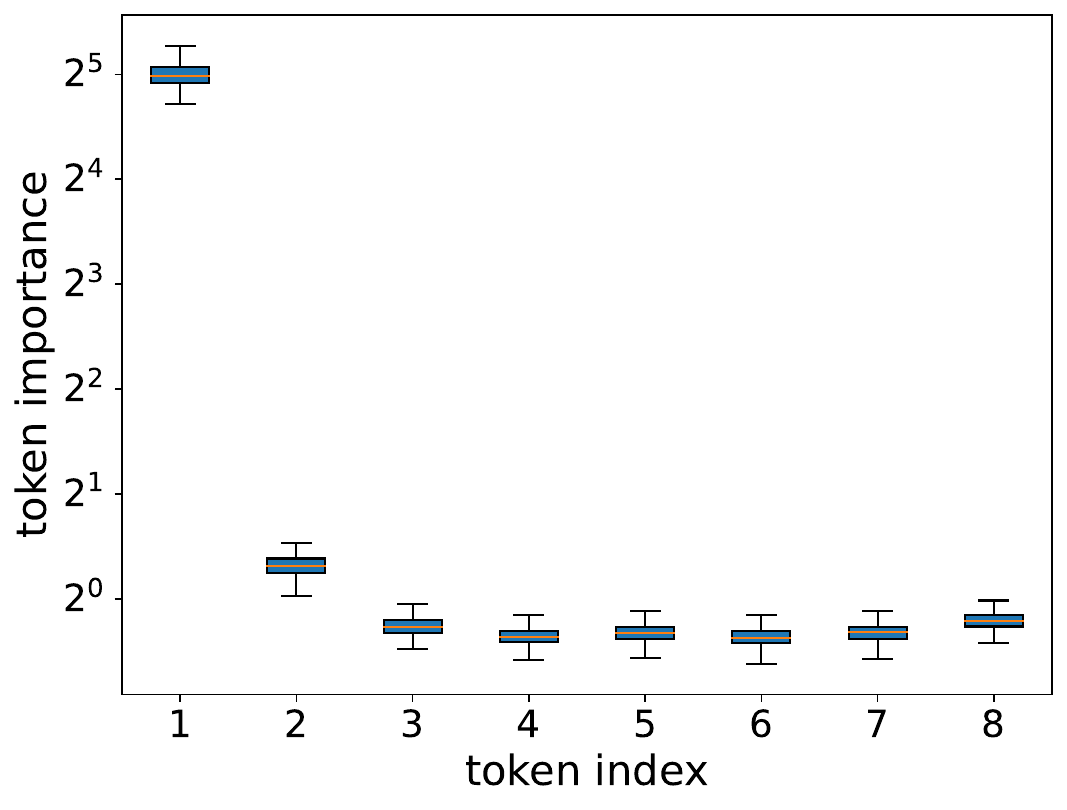}
        \caption{APH importance of top 8 tokens.}
        \label{fig:token_sub1}
    \end{subfigure}%
    \hfill
    \begin{subfigure}{.50\columnwidth}
        \centering
        \includegraphics[width=1\linewidth]{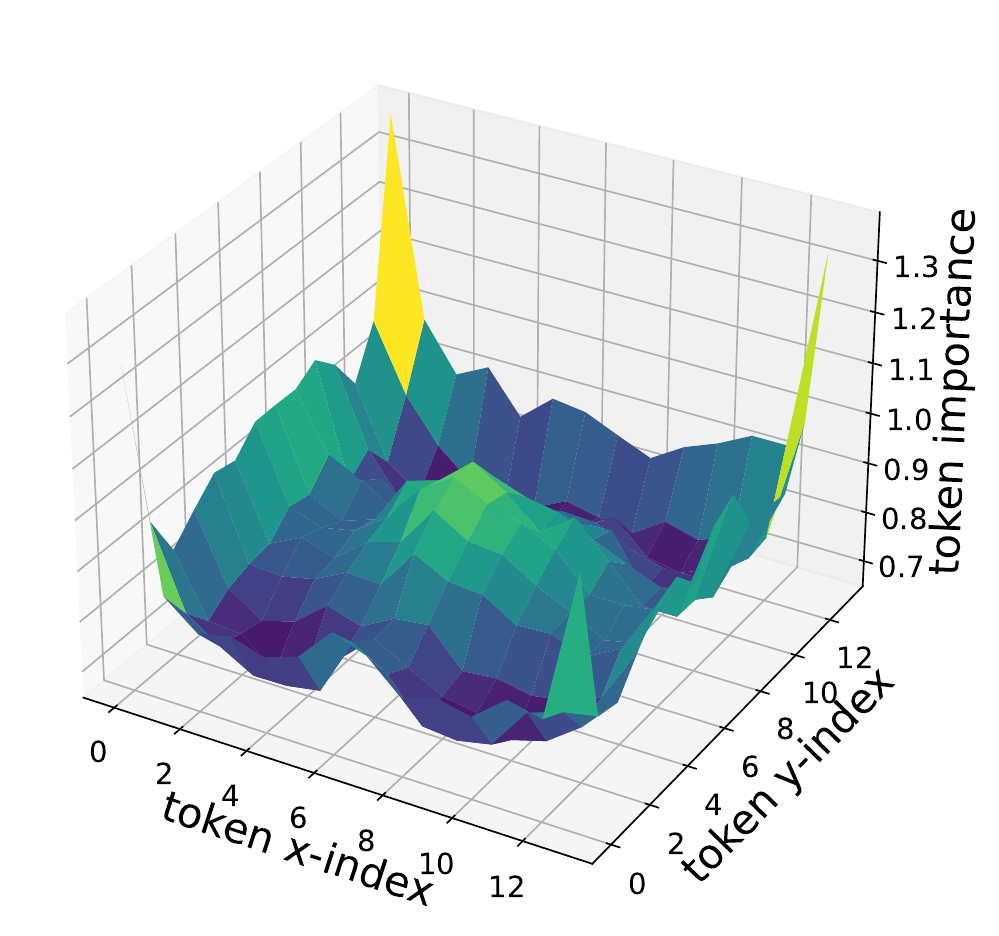}
        \caption{APH importance of patch tokens.}
        \label{fig:token_sub2}
    \end{subfigure}%

    \caption{Illustration on the token importance in \textit{ViT-S.blocks.7}.}
    \label{fig:token}
\end{figure}

\begin{figure}[t]
    \centering
    \includegraphics[width=1\linewidth]{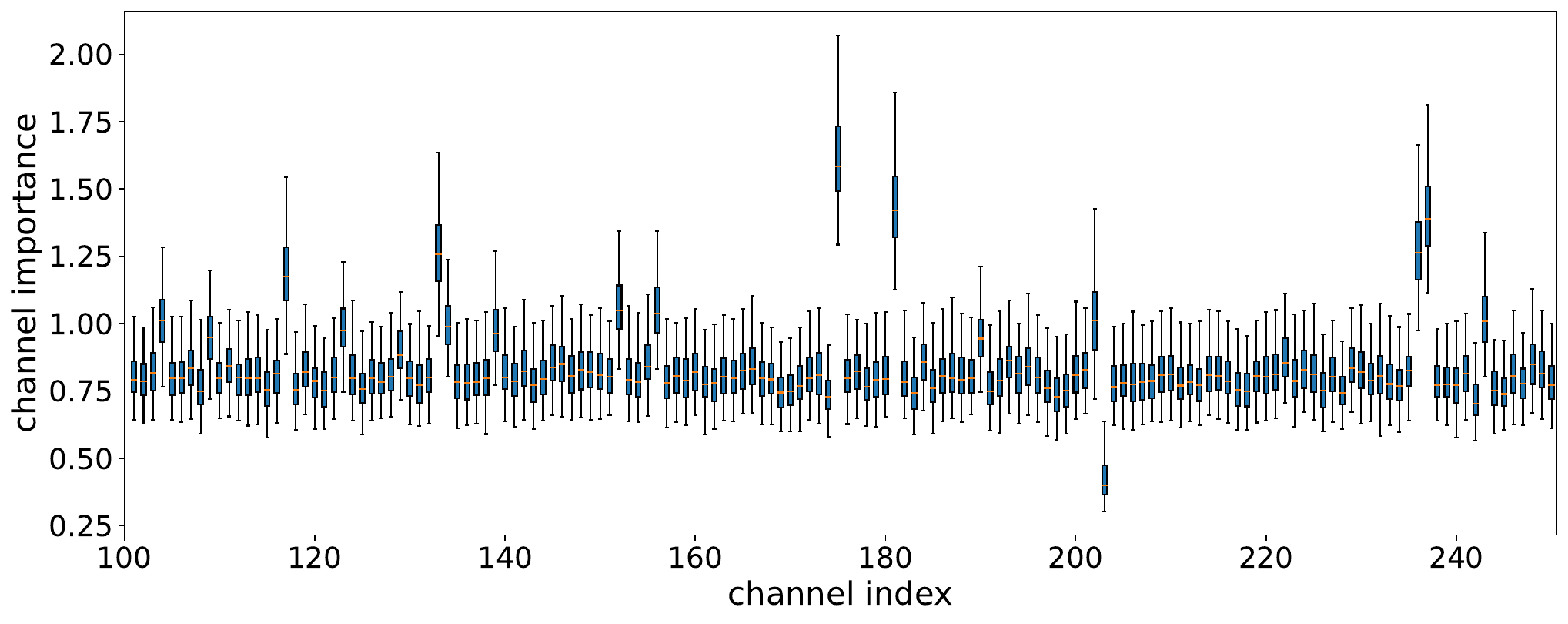}
    \caption{Illustration on the channel importance.}
    \label{fig:channel}
\end{figure}

\cref{fig:token} demonstrates the APH importance for tokens from \textit{ViT-S.blocks.7}, where \cref{fig:token} (a) displays the tokens with top 8 importance, and  \cref{fig:token} (b) shows the importance of the rearranged \(14\times 14\) patch tokens. It can be observed that the importance of the class token, the first one in  \cref{fig:token} (a), is much higher than that of the patch tokens, and distinct patch tokens have substantially different APH importance. Moreover, \cref{fig:channel} displays APH importance for the output channels with indices 100 to 250 from \textit{ViT-S.blocks.7}, indicating that the values of APH importance for certain channels are significantly higher than that of others.

The above visualization results indicate that the importance between distinct tokens or channels varies significantly in Vision Transformers, implying the necessity of incorporating important metrics during reconstruction.

\end{document}